\documentclass{article}
\usepackage{spconf}
\usepackage{subfigure}
\usepackage{amsmath,graphicx}
\usepackage{adjustbox}
\usepackage{amssymb,amsfonts}

\usepackage{algorithmic}
\usepackage{algorithm}
\usepackage{multirow}
\usepackage{url}

\usepackage{textcomp}
\usepackage{diagbox}

\graphicspath{{./pics/}}

\title{NetRCA: An Effective Network Fault Cause Localization Algorithm}
\name{Chaoli Zhang\textsuperscript{\rm 1}, 
Zhiqiang Zhou\textsuperscript{\rm 1}, 
Yingying Zhang\textsuperscript{\rm 2}, 
Linxiao Yang\textsuperscript{\rm 1}, 
Kai He\textsuperscript{\rm 3}, 
Qingsong Wen\textsuperscript{\rm 3}$^{\star}$\thanks{$^{\star}$All authors contributed equally. Qingsong Wen is the team captain.}, 
Liang Sun\textsuperscript{\rm 3}}
\address{
\textsuperscript{\rm 1}DAMO Academy, Alibaba Group, 
Hangzhou, China\\
\textsuperscript{\rm 2}Alibaba Cloud Intelligence, Alibaba Group, 
Hangzhou, China\\
\textsuperscript{\rm 3}DAMO Academy, Alibaba Group, 
Bellevue, USA\\
}

\providecommand{\keywords}[1]
{
  \small	
  \textbf{\textit{Keywords---}} #1
}

\begin{document}
%
\maketitle
\begin{abstract}
Localizing the root cause of network faults is crucial to network operation and maintenance. However, due to the complicated network architectures and wireless environments, as well as limited labeled data, accurately localizing the true root cause is challenging. In this paper, we propose a novel algorithm named NetRCA to deal with this problem. Firstly, we extract effective derived features from the original raw data by considering temporal, directional, attribution, and interaction characteristics. Secondly, we adopt multivariate time series similarity and label propagation to generate new training data from both labeled and unlabeled data to overcome the lack of labeled samples. Thirdly, we design an ensemble model which combines XGBoost, rule set learning, attribution model, and graph algorithm, to fully utilize all data information and enhance performance. Finally, experiments and analysis are conducted on the real-world dataset from ICASSP 2022 AIOps Challenge to demonstrate the superiority and effectiveness of our approach. 
\end{abstract}
%
%


\vspace{-3mm}
\section{Introduction}\vspace{-3mm}
The increase in size and complexity of networks call for automatic and intelligent root cause analysis algorithms and tools~\cite{asghar2018self,gonzalez2017root,sole2017survey}. 
Due to the diverse type of networks and complex key performance indicator (KPI) patterns of multivariate time-series data, developing robust and reliable fault localization solutions is challenging and has received lots of research attention.
In \cite{gomez2015automatic}, it proposes an automatic diagnosis system based on an unsupervised self-organizing maps. 
In \cite{mfula2017adaptive}, an automated fault detection and diagnosis solution called adaptive root cause analysis is designed, which uses measurements and other network data together with Bayesian network theory~\cite{cai2017bayesian} to perform automated evidence-based RCA.
In \cite{munoz2016root}, it proposes an automatic diagnosis algorithm to analyze the temporal evolution of a plurality of metrics and searches for potential interdependence under the presence of faults. 
In \cite{zhang2021cloudrca}, a root cause analysis system is designed via robust time series analysis~\cite{wen2021robustperiod,wen2019robuststl,yang2021robust} and hierarchical Bayesian network.
Recently, \cite{terra2020explainability} proposes the application of multiple global and local explainability methods with the main purpose of root cause analysis in networks by identifying important features contributing to the decision. 

Despite the aforementioned efforts, existing fault localization schemes remain difficult for complicated 5G networks due to three critical hurdles. The first challenge is that the increased depth of the network is likely to propagate errors through the causal path from the source nodes to the root node, making accurate root cause attribution difficult. The second challenge is the lack of adequate known labels. One usually has to dive deep into KPIs, service logs, and communication details before narrowing down the possible root cause candidates. Last but not least, the time-series data associated with each network node are multivariate. They are often in complex patterns with inter-dependence and noises,
resulting in difficulty extracting node relationships.



In this paper, we propose an effective wireless network fault cause localization algorithm called NetRCA to deal with these challenges based on the dataset from ICASSP 2022 AIOps Challenge in Communication Networks~\cite{ICASSP-SPGC2022}. NetRCA consists of three main components, including feature engineering, data augmentation, and model ensemble. In feature engineering, we designed features specifically for time series data and wireless direction related features. 
As labeled data are often limited while there exist large amount of unlabeled data in practice, we propose novel methods to perform data augmentation to generate labeled data. 
Lastly, we treat the root cause localization as a classification problem by using model ensemble, which not only adopts XGBoost to obtain a strong baseline but also leverages rule set learning, attribution model, and graph algorithm, to exploit the causal relationship graph for further performance enhancement.
Apart from its high prediction accuracy,
our NetRCA model is able to output interpretable results thanks to the adopted rule based models, which is helpful to understand
how the root cause affects the system.

\vspace{-3mm}
\section{Proposed Network RCA Framework}\vspace{-3mm}
\begin{figure*}[!t]
\centering
    \includegraphics[width=0.7\linewidth]{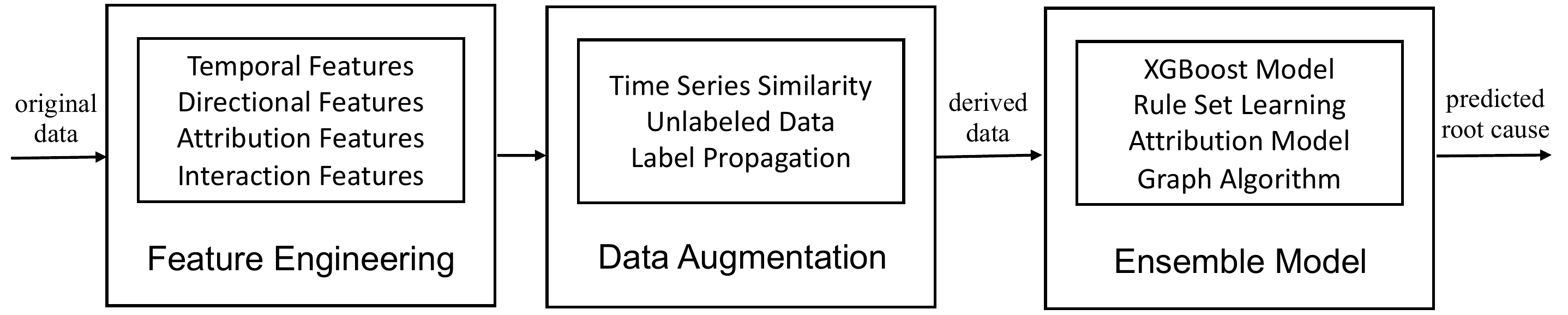}
    \vspace{-0.4cm}
    \caption{Framework of the proposed NetRCA algorithm.}
    \vspace{-0.6cm}
\label{fig_NetRCA}
\end{figure*}

\subsection{Framework Overview}\vspace{-2mm}
There are three main steps in the proposed NetRCA algorithm: 1) feature engineering, 2) data augmentation, 3) model ensemble. The framework of the NetRCA is depicted in Figure~\ref{fig_NetRCA}, which will be elaborated in the following sections.

\vspace{-4mm}
\subsection{Feature Engineering}\label{FE}\vspace{-2mm}
As the number of timestamps in each sample is different, directly training
the model using all the timestamps may lead to bias, i.e.,
the model is likely prone to focus the samples with more timestamp indices.
Thus, we train our model based on the features extracted from
each sample. The generated features can be roughly classified into 
four classes: temporal features, direction-related features, attribution features, and interaction features.

Some temporal features used in our model are based on
statistics of data, where the data in each timestamp are assumed to be
independent, including mean, minimum,
maximum, median, deciles, and skewness. We also include some
features to represent the shape of the time series, 
including the number of peaks and mean of change generated using
the public tsfresh\footnote{https://tsfresh.readthedocs.io/en/latest/index.html} package.


Multiple antenna and beamforming have been adopted in 5G network to enhance performance~\cite{li2021learning, shi2016sinr}. As mentioned in~\cite{ahmed2018survey,onggosanusi2018modular}, the direction of beamforming and the distances between each node play an essential role in network performance, which is also important for root localization especially between root cause 2 and 3 as indicated in the causal graph provided by the AIOps Challenge. We consider feature 20 an important feature in models for detecting root causes 2 and 3. Since feature 20 gives the index from 0 to 31 of each node which maps to a $4\times 8$ location matrix, we first transform each node's index into a two-dimensional coordinate, and then measure the distance between each pair of nodes via Euclidean distance. After that, we derive the features for capturing the interrelationship between feature X and Y, including feature 61/69/77/85 and feature 28/36/44/52, respectively, for further improvement. At last, statistical features (such as mean, variance, quantiles, etc.) are summarized from the distribution of the distances in each time slice sample for model training. 

We derive attribution features for all the nodes but the feature 0
according to the causal graph. As stated in the problem description,
these root cause finally leads to a lower value of feature 0.
Then the true root causes and their descendants will play a more
important rule on the current value of feature 0 than the others.
Thus, we generate a new feature as an
estimate of the importance score of each feature on 
predict feature 0, and detail derivation is
presented in Sec.\ref{pam}.


Two-order interaction features for $X$ and $Y$ are generated.
As feature $X$ equals the ratio of feature $Y$ over some unknown factors,
we generate feature $X$ over $Y$ to measure the effect of these 
unknown factors. Specially, we first group the features in $X$ and $Y$
to pairs according to the problem description. For each pair,
we compute the ratio of $X$ over $Y$.
Finally, we compute the statistics of these ratios as what
we do for temporal features.

\vspace{-4mm}
\subsection{Data Augmentation}\vspace{-2mm}

\textbf{Multivariate time series similarity:}
To measure similarity between multivariate time series with different lengths, we apply Eros (Extended Frobenius norm)~\cite{yang2004pca} algorithm to calculate the similarity. Eros extends the Frobenius norm by using the principal components and computing the similarities based on eigenvectors.
Formally, let $\mathbf{A}$ and $\mathbf{B}$ be two MTS of size $m_A\times n$ and $m_B\times n$, respectively. Let $\mathbf{V}_A=[a_1, \cdots, a_n]\in\mathcal{R}^{n\times n}$ and $\mathbf{V}_B=[b_1, \cdots, b_n]\in\mathcal{R}^{n\times n}$ be two right eigenvector matrices by applying SVD to the covariance matrices of $\mathbf{A}$ and $\mathbf{B}$, respectively. Then the Eros similarity of $\mathbf{A}$ and $\mathbf{B}$ is
\vspace{-1mm}
\begin{equation}
    \text{Eros}(\mathbf{A}, \mathbf{B}, w) = \sum_{i=1}^n w_i |<a_i, b_i>|,\vspace{-1mm}
\end{equation}
where $w=[w_1, \cdots ,w_n]$ is the weight vector based on the eigenvalues and it satisfying $\sum_{i=1}^n w_i=1$.

\vspace{1mm}
\noindent\textbf{Data/Label augmentation:} Data augmentation is important for learning from time series as the labeled data is often limited~\cite{wen2020time}. We find more than half of the training data are unlabeled, and simply dropping these data would miss a lot of valuable information. Using Eros, we are able to measure the similarities between any two training samples, both of which are multivariate time series. In this way, we can enrich the training set by selecting the samples from those unlabeled data with high similarities to the labeled data, and label them according to the true root causes of those training samples similar to them. This process is executed for each type of the root causes respectively to improve the calculation efficiency.

Another important augmentation is propagating root cause labels of training samples that share similar timestamps. This improves the predictions of multi-root causes for the test dataset, especially for root cause 1. Almost all training samples labeled as root cause 1 are not associated with the other two root causes, so it seems reasonable to assume root cause 1 is more likely to occur independently of the remaining root causes. However, taking a close look at the timestamps and labels of all training samples, it is evident that there are plenty of one-minute time intervals where the root cause 1 co-occurs with root causes 2 and 3. Missing propagating root cause 1 label to samples where it highly likely exists tends to jeopardize the quality of supervised learning. As a result, we have aligned all training samples by their timestamps and augmented their true labels as the union set of all root causes labels.

\vspace{-4mm}
\subsection{Ensemble Model}\label{rule}\label{pam}\label{graph}
\vspace{-2mm}
The NetRCA adopts ensemble model to predict root cause, which applies XGBoost to obtain initial outcomes and then combine rule set learning, attribution model, and graph algorithm to refine the outcomes for final results. The details are described as follows.

\vspace{1mm}
\noindent
\textbf{Root Cause Classification via XGBoost:} In our solution, we treat finding the correct root cause as a classification problem. Specifically, we apply XGBoost~\cite{XGBoost} as our base model due to its good performance.
Note that there exists the problem of unbalanced labels for different roots, so in our model we adjust the balance of positive and negative weights for better results.


\vspace{1mm}
\noindent\textbf{Rule Set Learning:} One challenge of building a powerful classifier is the feature interaction, which occurs when the values of some features influence each other. The presence of feature interaction makes the output cannot be represented as a summation of the effects of individual features. Decision rule set~\cite{Dash:boolean:rule}, which consists of a set of ``IF ... THEN ...'' logical rules, can handle the feature interaction naturally. Specifically, a rule, which is the logical conjunction of clauses (comparison of features and thresholds), builds a logical relationship between the target and the features, and is capable of modeling the nonlinear interaction of features and target. Another important property of the rule set is its interpretability. The logical structure of the rules makes them easy to interpret. The interpretability of the rules enables us to understand the relationship between features and target, and helps to detect import features. We use the public Skope-rules package\footnote{https://github.com/scikit-learn-contrib/skope-rules} to learn rules from data. Skope-rules generate rule candidates using tree models. They build a number of decision trees and treat a path from the root node to an internal node or leaf node as a rule candidate. These candidates are then filtered by some predefined criteria such as precision and recall. Only those with precision and recall above their threshold are remained. Finally, the similarity filtering is applied to select rules with enough diversity. In our solution, we apply Skope-rules to learn the potential rules for each root cause and 
drop the rules that predict a sample not belongs to any cause. 

\vspace{1mm}
\noindent\textbf{Predictive Attribution Model:} When the interdependent relationship between nodes is available, it is possible to estimate the importance of the features. Feature importance measures the marginal gain of adding a particular feature to the causal graph. Intuitively, anomaly data in upstream nodes will likely
contribute a large portion of changes to feature 0, which can help us identify the root cause. To this end, we generate a new feature
that measures the feature importance for each sample and integrate them into our model. Our feature importance estimation is based on the Shapley value~\cite{shapley}. Given a set of features $S$, the relationship $f$ between
internal and feature 0, and let $\boldsymbol{x}_T$ be the subset of $\boldsymbol{x}$ that only contains the features in $T$, the Shapely value $\phi(i)$ of the feature $i$ is
\begin{align}
\phi(i)=\sum_{T\subseteq S\setminus\{i\}}\frac{|T|!(p-|T|-1)!}{p!}(f(\boldsymbol{x}_{T\cup\{i\}})-f(\boldsymbol{x}_T)),\vspace{-1mm}
\end{align}
which measures the average marginal gain that adds the feature $i$ in different orders. Nevertheless, directly computing the Shapley value raises two difficulties. Firstly, the function $f$ only produces an output when all the features are ready, and one cannot estimate the output of $f$ only given part of the features. Secondly, computing the Shapley value is time-consuming, as it requires computing the marginal gain of all the possible orders. To address these issues, we approximate $f(\boldsymbol{x}_T)$ using $f(\boldsymbol{x}_T,\boldsymbol{\bar{x}}_{S\setminus T})$, where $\bar{x}_i$ denotes the average value of the feature $i$. In other words, $f(\boldsymbol{x}_T)$ is approximated as the output of $f$ on input $[\boldsymbol{x}_T,\boldsymbol{\bar{x}}_{S\setminus T}]$, where we keep the 
features in $T$ unchanged and set the remain features to their mean value.
We note that this is a common strategy used for computing the Shapley values. To overcome the second difficulty,
we approximate Shapley value as the value reduction of $f$ when remove $i$
from $S$, i.e.\vspace{-1mm}
\begin{align}
\phi(i)\approx|f(\boldsymbol{x}_S)-f([\boldsymbol{x}_{S\setminus\{i\}},\bar{x}_i])|\vspace{-1mm}
\end{align}
We note that such an approximation works well, especially for sparse causal graph. 
In our experiment, we estimate the relationship function $f$ between
internal and feature 0 nodes by training a XGboost model.
After estimate the feature importance,
we identify the root cause by simply comparing it with a pre-defined
threshold. Those root causes with an importance higher than threshold
are identified as the true root cause.


\vspace{1mm}
\noindent\textbf{Graph Algorithm:} To further exploit the provided causal graph, we design a specialized graph algorithm based on univariate time series similarity to rank and locate the true root cause. 
The first motivation is that the features next to the root cause (e.g., feature13 and feature15 for root 1) should show a high correlation in a similarity metric to the target feature 0. 
Since feature 0 is the target variable that the operator cares about and values of the features/KPIs vary with time and affect each other, we compute the absolute value of Pearson correlation as a similarity score $S_i$ between feature $i$ and feature 0 as
\begin{equation}
    S_i = \left|\frac{\sum_{t=1}^T([\mathbf{f}_i]_t - \bar{\mathbf{f}}_i)([\mathbf{f}_0]_t - \bar{\mathbf{f}}_0) } {\sqrt{\sum_{t=1}^T([\mathbf{f}_i]_t - \bar{\mathbf{f}}_i)^2} \sqrt{\sum_{t=1}^T([\mathbf{f}_0]_t - \bar{\mathbf{f}}_0)^2 } }\right|,\vspace{-1mm}
\end{equation}
where $\mathbf{f}_i$ is the univariate time series data of feature $i$ and $\bar{\mathbf{f}}_i$ denotes its mean value.
Note that before calculating the Pearson correlation, we linearly interpolate the missing data of all features. The Pearson correlation measures how two features (time series) co-vary over time and indicate the relationship from 0 (not correlated) to 1 (perfectly positively/negatively correlated). This correlation-based similarity score $S_i$  signifies the relevance of the feature $i$ to the target feature 0.
Since correlation does not always imply causality, using a similarity score may result in false positives. Instead, we consider both similarity score and causal relationship graph to enhance performance as our second motivation. Specifically, we adopt the popular graph algorithm, personalized PageRank~\cite{jeh2003scaling}, to exploit the causal graph. The main idea is to conduct a random walk over the causal graph based on similarity scores. Specifically, starting from feature 0, the features are selected in sequence by randomly picking up the neighbor feature in the causal graph. The pickup probability is proportional to the edge weights which are calculated by normalized similarity score as 
$w_{ij} = {A_{ij}S_j}/{\sum_j A_{ij}S_j}$, 
where $A_{ij}$ is 1 if the two features $\mathbf{f}_i$ and $\mathbf{f}_j$ are connected and 0 otherwise. The final root is based on the assumption that the more visits to the feature next to the root cause (e.g., feature 13, 15, 19, or 60), the more possible that root cause is the true root of feature 0.


\vspace{-2mm}
\section{Experiments and Discussions}\vspace{-2mm}
In this section we summarize and discuss the performance of the NetRCA on the dataset of ICASSP 2022 AIOps Challenge~\cite{ICASSP-SPGC2022}.
\vspace{-3mm}
\subsection{Datasets and Evaluation Metrics}\vspace{-2mm}

The Challenge datasets include one fixed causal relationship graph and feature dataset containing 2984 samples with 23 observable variables. 
Among the 2984 samples, only about 45\% of them are labeled with root cause faults while others remain unlabeled, which indicates labels are scarce and incomprehensive. 

For the evaluation metric, we adopt the normalized final score provided by the Challenge, which increasing 1 mark for each true positive root while deducting 1 mark for each false positive root. The final mark is normalized by the number of test samples, so the highest final score would be 1.


\vspace{-3mm}
\subsection{Implementation and configuration}\vspace{-2mm}

As discussed in section~\ref{FE}, we generate various features from the raw data. However, with limited number of training samples, just training model on all these features leads to model overfitting. We need to conduct feature engineering and selection carefully. 
At the very beginning, we tried multi-class classification models with class label set \{root1, root2, root3, root2\&root3\} from train data. However, with such setting, we struggled with score 0.7+. Such model has several shortcomings. First of all, all features are simply shared among root cause 1, root cause 2 and root cause 3. It is not necessary to add feature 20s or feature X, feature Y when root cause 1 is considered. Secondly, with label set \{root1, root2, root3, root2\&root3\}, an implicit statement is that root cause 1 and root cause 2 won't appear simultaneously, nor root cause 1 \& root cause 3 or root cause 1 \& root cause 2 \& root cause 3. In reality, such assumption is limited and not general.

Based on the above observations and experimental results, we turn to train three binary classification models for root cause 1, root cause 2 and root cause 3 respectively. Based on the derived data from feature engineering and data augmentation, for the model of root cause 1, we mainly use the information from feature 0, 13, 15 and the interaction among them. With features set for root cause 1 changed from \{0, 13, 15, 19, 20, X, Y\} to \{0, 13, 15\}, the test score increased from 0.825 to 0.837 with the same parameters, which verifies the effectiveness of the three binary classification models. For the model of root cause 2, the main used features capturing useful information are from feature 19 and 20. For the model of root cause 3, we use the generated feature for capturing information in feature X and Y, as feature Y contains significant information to distinguish root cause 3 from root cause 2. 


Besides, given the provided causal graph, it can be seen that different root causes are related with different features. To enhance the overall performance, we adopt ensemble modeling by first training XGBoost models for root1, root2 and root3 separately with different features as discussed above, and then the results are further enhanced by rule set model, attribution model, and graph model as described in section~\ref{rule}. 
Next, we will discuss insights from the ensemble models, as well as ablation studies to demonstrate the performance improvement of our final NetRCA.


\begin{figure}[!t]
    \includegraphics[width=1.0\linewidth]{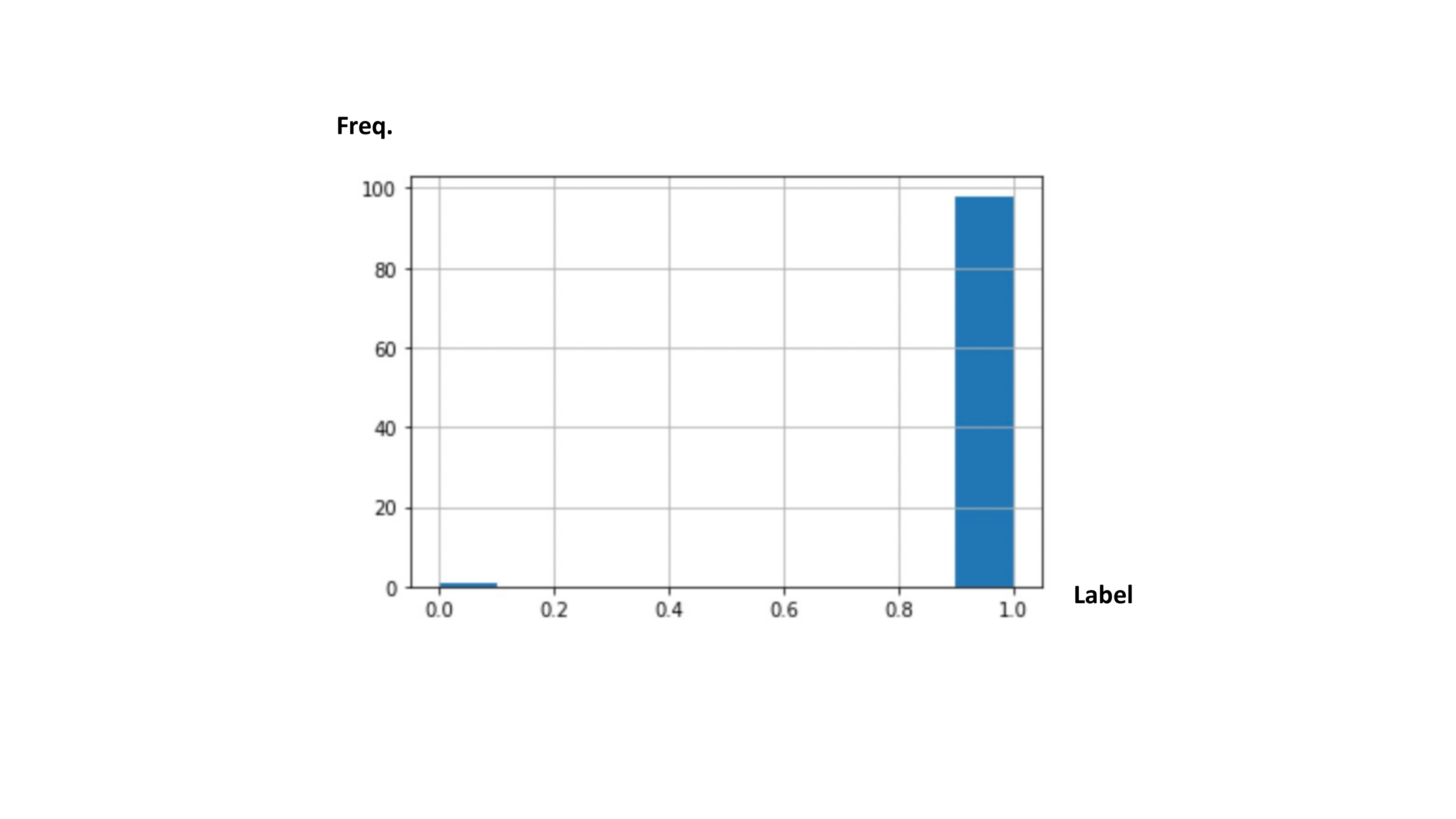}
    \vspace{-19mm}
    \caption{Histogram of samples covered by rule "$feature13_{min} \le 1.75e^5$ and $feature13_{max} \le 4.00e^5$ and $feature13_{quantile0.4} \le 1.92e^5$" to predict root 1.}
\label{fig:rule-example}\vspace{-5mm}
\end{figure}
\vspace{-4mm}
\subsection{Model Interpretability}\vspace{-2mm}
In this subsection, we demonstrate the effectiveness of interpretability in helping diagnose the model performance and improve human trustiness. In Figure~\ref{fig:rule-example}, we show a histogram of samples covered by one of the interpretable rules generated to predict root cause 1. It is evident that the accuracy of the rule is close to one. More importantly, the boolean rule depicted in the caption implies that a lower value of feature 13 (including min, max, and quantile distributions) is more likely linked to root cause 1. The causal graph confirmed that root cause 1 is often associated with insufficient resources tied to features 13 and 15, which matches the intuition behind the rules. By applying the interpretable models proposed by NetRCA, one can develop a deeper understanding of how the model makes predictions and correct the issues of predicting right answers for wrong causes.

\begin{table}[!t]
\centering
\caption{Ablation studies of the proposed NetRCA model.}
\scalebox{0.8}{
    \begin{tabular}{c| c c c |c }
    \hline
     Models & \!Root1 acc\! & \!Root2 acc\!  & \!Root3 acc\! &\!Final Score\!\\
    \hline\hline
   XGB & 0.9828 & 0.97849 & 0.9957  & 0.78139 \\\hline
    XGB+FE & 0.9957 & 0.97849 & 0.9914 & 0.86611 \\\hline
    XGB+FE+Graph & 0.9957 & 0.97849 & 0.9914 & 0.87917 \\\hline
  \textbf{Proposed NetRCA} & 0.9957 & 0.98495 & 0.9914 & \textbf{0.91778}  \\\hline
    \end{tabular}
}
    \vspace{-5mm}
    \label{Ablation}
\end{table}


\vspace{-4mm}
\subsection{Performance Comparison and Ablation Studies}\vspace{-2mm}

For ablation studies, we first split the 1407 labeled samples into training set and validation set with size 942 and 465, respectively. In Table~\ref{Ablation}, we compare the baseline XGBoost model without any extra features (XGB), the XGBoost model with features (XGB+FE) generated in feature engineering as described in section~\ref{FE}, the combination of XGB+FE and graph algorithm (XGB+FE+Graph), and the NetRCA algorithm which combines XGB+FE+Graph with data augmentation, rule set learning and attribution model.
Specifically, the first 3 columns of Table~\ref{Ablation} denote the accuracy of the model's performance on the validation set belonging to each root cause (root cause 1, 2 and 3), while the forth column indicates the submitted score for generated solution on the test data.

There are several key insights we can get from the results in Table~\ref{Ablation}: 1) All the models even the basic XGB model can achieve excellent accuracy in the training data. However, the submitted score demonstrate that there exist certain gap between the distributions of training and testing data. The three ablated models show overfitting in different degrees, while our NetRCA algorithm would prevent overfitting and give us a more robust solution. 2) XGB+FE model outperforms basic XGB significantly in both training and validation set thanks to feature engineering described in section~\ref{FE}. By diving into the business background and extracting effective information from temporal features,  direction-related features, attribution features, and interaction features, we are able to obtain an complete perspective and discover some underlying rules. 3) Although there is no remarkable improvement in the training set, combining graph model can increase the final submission score by more than 1\%. The reason may lie in that the graph model can help us better capture the causal relationships among those features. 4) Both feature engineering and graph model seem have no influence on the accuracy of root cause 2 on the training set, since the samples belong to the root cause 2 are very limited. But our final NetRCA shows a noteworthy enhancement in the accuracy for root cause 2 due to the data augmentation which can resolve the problem of imbalanced data. Besides, another challenge for identifying root cause 2 lie in the concurrence of root cause 2 and other root causes. In this way, rule set learning and attribution model can reduce the mutual influence among those features, and further improve the final score.


\vspace{-2mm}
\section{Conclusions}\vspace{-2mm}
In this paper we propose a novel algorithm named NetRCA to localize the root cause of network faults. Besides carefully designed feature engineering, our algorithm adopts data augmentation to generate new training data to overcome the lack of labeled samples.
Furthermore, we design an ensemble approach which effectively combines different models 
to perform accurate and reliable causal inference for the network faults.


\vfill\pagebreak

\clearpage
\bibliographystyle{IEEEbib.bst}
\bibliography{6_bibfile}

\end{document}